%% file: main.tex
\documentclass{article}

\usepackage{PRIMEarxiv}

\usepackage[utf8]{inputenc} 
\usepackage[T1]{fontenc}    
\usepackage{hyperref}       
\usepackage{url}            
\usepackage{booktabs}       
\usepackage{amsfonts}       
\usepackage{nicefrac}       
\usepackage{microtype}      
\usepackage{lipsum}
\usepackage{graphicx}
\graphicspath{{media/}}     

\title{Can AI Autonomously Build, Operate, and Use the Entire Data Stack?}

\author{
  Arvind Agarwal, Lisa Amini, Sameep Mehta, Horst Samulowitz, Kavitha Srinivas\\
  IBM Research\\
  \texttt{\{arvagarw, sameepmehta\}@in.ibm.com, \{lisa.amini, hsamulo\}@us.ibm.com}, kavitha.srinivas@ibm.com \\
}

\begin{document}

\maketitle

\begin{abstract}
\input{body/abstract_new}
\end{abstract}



\section{Introduction}
\input{body/intro_new2}

\section{The Data Stack}
\label{sec:datastack}
\input{body/datastack_new}

\section{Towards realizing an Autonomous Data Stack}
\input{body/vision_new}

\section{Opportunities in an Era of Autonomous Data Stacks}
\input{body/data_stack_opportunities}


\section{Conclusion}
\input{body/conclusion}

\section{Acknowledgments}

\input{body/acknowledgements}

\bibliographystyle{unsrt}  
\bibliography{data_stack}

\end{document}

%% file: body/abstract_new.tex
Enterprise data management is a monumental task. It spans data architecture and systems, integration, quality, governance, and continuous improvement. While AI assistants can help specific persona, such as data engineers and stewards, to navigate and configure the data stack, they fall far short of full automation. However, as AI becomes increasingly capable of tackling tasks that have previously resisted automation due to inherent complexities, we believe there is an imminent opportunity to target fully autonomous data estates.

Currently, AI is used in different parts of the data stack, but in this paper, we argue for a paradigm shift from the use of AI in independent data component operations towards a more holistic and autonomous handling of the entire data lifecycle.
Towards that end, we explore how each stage of the modern data stack can be autonomously managed by intelligent agents to build self-sufficient systems that can be used not only by human end-users, but also by AI itself.  We begin by describing the mounting forces and opportunities that demand this paradigm shift, examine how agents can streamline the data lifecycle, and highlight open questions and areas where additional research is needed. We hope this work will inspire lively debate, stimulate further research, motivate collaborative approaches, and facilitate a more autonomous future for data systems.







%% file: body/intro_new2.tex
In the evolving landscape of data-driven decision-making, automation of the data stack can become a critical enabler for scalable and efficient enterprise operations. The need for such automation has never been greater, given that modern enterprises operate in an increasingly data-saturated environment, and success hinges on the ability to extract insights rapidly, securely, and cost-effectively from ever growing and diverse datasets. Modern data stacks enabling these insights and decision making have grown beyond the practical limits of traditional, human-centric approaches, making the automation mandatory. While automation helps in doing current tasks more efficiently and scalably, we propose that the bigger benefits will be realized when we move from automation to autonomy.

The rapid advancement of LLMs and autonomous agents is transforming every facet of data management. Today’s agents can seamlessly orchestrate tools to ingest, transform, and enrich data, while also enabling users to interact with complex systems through natural language and multimodal interfaces \cite{li2024can, lei2024spider, Ma2025AutoData}. However, we argue that current efforts remain predominantly focused on streamlining predefined tasks rather than achieving true autonomy, where agents can collaborate, reason, adapt, and make decisions in dynamic data environments. In this work, we contend that the data management stack is on the cusp of a foundational shift driven by what we term \textbf{Agentic DataOps}, where intelligent agents play a central, decision-making role across the entire data lifecycle.


To motivate our point of view, let us go over a use case. Consider a simplified yet representative scenario in financial analytics: 

\begin{quote}
\textit{ Create a financial product that forecasts performance across Asian and European mutual funds. Additional requirements include low latency, GDPR compliance, price-performance, high data quality, restricted user access.}
\end{quote}

Achieving this  straightforward goal with today’s state-of-the-art systems involves a complex, multi-stakeholder process. A data steward is responsible for dataset acquisition, registration, lineage, and profiling. A data engineer must provision and orchestrate data pipelines. A site reliability engineer (SRE) ensures system availability and latency compliance, and a chief information security officer (CISO) must review and enforce access controls. A data scientist designs and evaluates forecasting models. These roles operate largely in silos, relying on manual coordination, customized automation scripts, and piecemeal governance workflows. As a result, the end-to-end process, which should have been completed within hours—often stretches into days or weeks. The situation becomes even more complex when issues with latency, data quality, security, or pipeline failures arise.  Coordination is required to identify the root causes and apply fixes, and finally, to build a data management system that is effective, efficient and long lasting; which brings up a fundamental question:

\begin{quote}
\textit{Could artificial intelligence (AI) build the entire system, ranging from data infrastructure design, to data  discovery, integration, governance and insights, and therefore the entire data stack from scratch?}
\end{quote}

\begin{figure*}
    \centering
    \includegraphics[width=\textwidth, trim={0.9cm 1.0cm 1.0cm 3.5cm},clip]{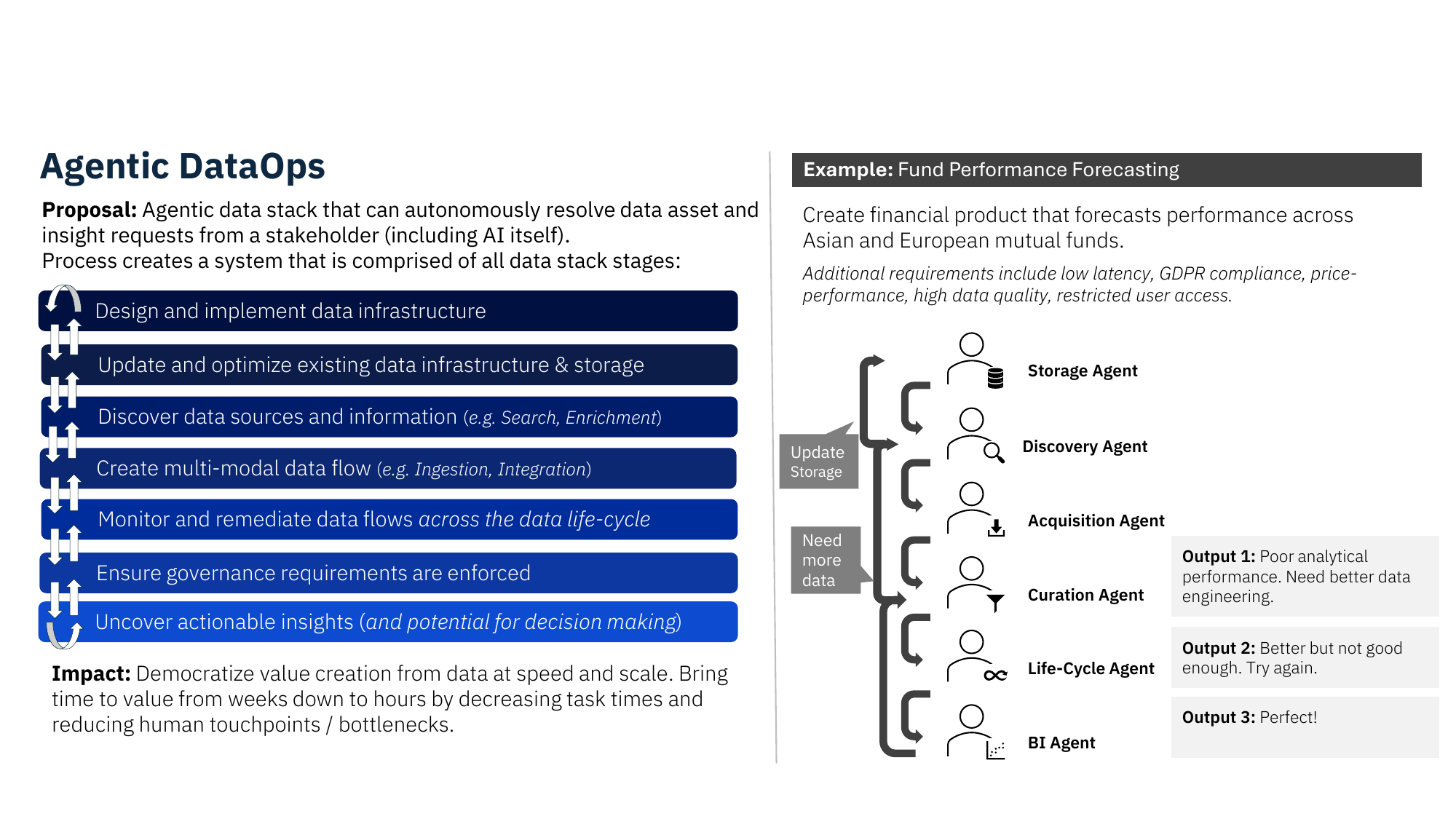}
\caption{Illustration of the Agentic DataOps approach, detailing its core processes for an autonomous data stack and an example application in fund performance forecasting using specialized agents. \textit{Left:} A high-level summary of the main stages in a data stack starting with designing and building the storage infrastructure to generating insights. \textit{Right:} Example of changes moving back and forth across the entire stack. For instance, based on poor analytical performance, more data engineering is applied on the underlying data. However, the resulting outcome is still not satisfactory, and consequently additional data sources are added, which in turn requires updating the storage infrastructure and subsequently triggers the data ingestion and processing steps, before a new analytical model is built.}
  \label{fig:ADS}
\end{figure*}

Towards this goal, we propose an Agentic Dataops system where the only human intervention is for clarification of goals, approval for state-changing actions, and receiving results. All other tasks are performed autonomously through collaborating agents. The agents configure and interact with subsystems such as databases, lakehouses, and catalogs, as needed. To illustrate this, we revisit the example of the financial analytics system to explore how agents could function autonomously or semi-autonomously across the entire data stack. Figure~\ref{fig:ADS} provides an abstract view of the data stack (on the left) from designing and implementing the data infrastructure to uncovering actionable insights, and, on the right, agents working across the stack to service the financial analytics system request. Arguably, one could refine or combine the shown stages in different ways, we see this as an abstraction that enables us for this broader analysis.

In this scenario, a natural language interface allows the analyst to express the request through a multi-turn conversational interaction. The system processes this input by invoking relevant agents based on data availability and task complexity, including (but not limited to) agents to design and provision infrastructure, discover and acquire relevant data, assess and remediate data quality issues, monitor data lifecycle, diagnosis and remediate runtime issues, generate and visualize forecasts, leverage feedback signals to optimize performance. The goal is for these agents to collaborate—autonomously or with minimal human oversight—to fulfill the request. Security and compliance can be managed by embedding CISO agents within the pipeline, depending on regulatory requirements. A key strength of such an architecture lies in its scalability as well as extensibility: the agentic system is fully customizable through modular tooling and can be adapted to specific enterprise needs.

In the remainder of this paper, we investigate this question more deeply and examine how agentic systems can streamline the full data lifecycle—from infrastructure design to insight generation—and how each stage must evolve to support increasingly complex use cases and business demands. At the core, we assess whether artificial intelligence can architect, manage, and govern the entire data stack autonomously, potentially enabling AI to build complete data-driven systems—and, by extension, entire businesses—from the ground up. For each major component of the modern data stack, we analyze where AI is positioned to drive transformative change and where its impact is likely to be limited or incremental.

%% file: body/datastack_new.tex
We use the abstract view of the data stack from Figure~\ref{fig:ADS} to discuss the potential as well as challenges for AI infusion. The agents illustrated are not mapped directly to human persona, such as data architects, engineers, administrators, modelers, and stewards, but instead to tasks to be performed, such as data infrastructure design or data acquisition. Likewise, since, depending on the size of an enterprise, a single individual may serve in multiple roles, for example, data architect, engineer and administrator, we will use the term "data engineer" broadly to refer to those responsible for enterprise data management and operations.

\subsection{Designing and Implementing Data Infrastructure}
Traditionally, data engineers are responsible for selecting appropriate databases, designing schemas, and provisioning infrastructure. For financial asset prediction, this might involve setting up a DB2 or PostgreSQL database for structured data and a vector database for unstructured documents like U.S. Securities and Exchange Commission (SEC) filings. In addition, creating data pipelines results in setting up tools such as Airflow and governance might require a data catalog such as Unity. This process is time-consuming and requires domain expertise. With automation, an agent could analyze the use case and make an appropriate selection. Actual choices of engines needs to be based on query workloads, infrastructure costs, legal compliance to software, and access control requirements. It could also generate initial schemas, provision cloud infrastructure obeying resource requirements with minimal human intervention, generate code to provide data access and so on, all of which are an opportunity from increasing AI capabilities. Related work in this space includes LLM based schema design based on natural language specifications \cite{wang2025text2schemafillinggapdesigning}, and schema discovery and refinement based on files in a data lake \cite{10.14778/3685800.3685897}.

\textbf{AI Challenges:}
Gathering information about different backend engines to help plan the schema and design can be challenging because query latency is a big factor in the choices here, and query latencies are engine specific, which means the process must be iterative - one needs to build `test' schemas to test the ideas on different engines, and then choose based on generated queries. Within the underlying data infrastructure, substantial systems-level challenges persist—not only to support agentic exploration~\cite{xu2025systemsfoundationsagenticexploration}, but also to enable autonomous data stacks that can dynamically evolve and reconfigure in response to changing conditions during real-world deployments while still remaining efficient.
Cost is a another significant factor for generative AI models, which is motivating some practitioners to increasingly use symbolic and cost-efficient tools.
We expect solutions will eventually need to rely on both neural and symbolic systems to achieve automation; building and testing the efficacy of such systems is an area for future work.

While AI agents are likely capable of applying basic data security during this design and implementation phase, guaranteeing data security autonomously appears to be especially challenging given for instance active external actors trying to undermine security with ever evolving techniques.  

\subsection{Updating and Optimizing Data Infrastructure}
As data and demand grows, infrastructure must be updated and optimized. Engineers must monitor the data infrastructure to ensure it operates efficiently, scales effectively, and addresses security and compliance. When issues are detected or predicted, engineers may tackle by updating server, storage or network configurations, refactoring schemas, adding indices, partitioning workloads, or other strategies. Deciding which strategies to employ to minimize downtime during transitions can be challenging.

An agent can proactively monitor system performance, identify and diagnose issues and propose remediations, such as: schema optimizations, which resources to scale, and tradeoff explanations for alternative approaches. It can also apply indexing and partitioning strategies to improve query performance as workloads evolve, ensuring the system remains responsive as data volume increases. Relevant research in this area includes learned DBMS components, automated database tuning~\cite{10.1145/3709652, Kanellis2022LlamaTuneSD, zhao2025neurdb, 10.14778/3685800.3685838, 10.1145/3733620.3733641}, foundation database models~\cite{wehrstein2025foundation}, LLM inspired database cost modeling~\cite{ozcan2024symphony} and use of LLMs for network designs~\cite{hamadanian2025gliahumaninspiredaiautomated}. 

\textbf{AI Challenges:}
Agents will need to take into account the entire workload and not just a given query. This is needed to ensure performance in one area is not degraded while improving in another. Also, some performance issues may be buried deep inside application code. Unless the organization is rigorous about application logging, it will be difficult for AI to troubleshoot issues. We anticipate the need for better and more controlled pipelines and logging, in order to make diagnosis and remediation feasible.  Additionally, evaluating alternatives and assessing unintended consequences of updates is challenging.  AI would need to not only identify potential remediations, but also generate simulation and test environments in order to evaluate alternatives.  Building such simulation environments is an opportunity for future research.

\subsection{Discovering and Enriching Data Sources}
Identifying and integrating new data sources is often a manual task involving research and custom integration. For financial predictions, this includes U.S. Securities and Exchange Commission (SEC) filings, financial APIs, event databases, and news feeds.

Agents can automate this by searching for and cataloging relevant APIs and datasets, scraping web sources, and enriching data with metadata such as sentiment scores or company identifiers. They can also continuously update the data source catalog as new information becomes available. Furthermore, different schemas from diverse sources needs to be aligned and mapped to create a common semantic data model, which may also need to expand as data objects increase. Enrichment and mapping are complex tasks where LLMs have shown initial promise~\cite{DBLP:conf/semtab/2024, parciak2024schemamatchinglargelanguage}. Similarly, various LLM based approaches exist for discovery of structured data~\cite{majumder2024discoverybenchdatadrivendiscoverylarge,dataproductsearch, ledd}. 
Discovery and extraction of structure from unstructured data in multiple modalities is currently the focus of work in AI~\cite{docling,shankar2024docetl}, as is the integration of data, across silos.

\textbf{AI Challenges:}
Metadata extraction and enrichment based on feedback is a continuous task, especially for widely used applications where such feedback is often available. Effective management of such incremental data while ensuring its backward compatibility remains an active area of research. For instance, gradual incremental indexing (from coarse-grained to fine grained) can be used to ensure effectiveness and costs in check, particularly for large scale and often expensive LLMs. 
Discovery across large-scale, heterogeneous datasets remains a significant challenge. A key issue, as highlighted in recent work~\cite{lakediscovery}, is devising representations that capture sufficient semantic and structural information to enable efficient and accurate discovery across both structured and unstructured data sources.

\subsection{Creating Data Flows: Ingestion, Integration, Quality}

There are growing efforts to automate the generation of ETL/ELT pipelines using schema definitions, source metadata, and natural language directives~\cite{ELT-Bench,shankar2024docetl}. Recently, also AutoData~\cite{Ma2025AutoData} proposes a novel multi-agent system that automates the collection of datasets from open web sources using natural language instructions.
These pipelines can integrate with existing tools to monitor data, detect anomalies, and apply predefined quality rules and transformations. 
In the context of our use case, this enables seamless integration of diverse data sources—both structured and unstructured.

\textbf{AI Challenges:}
While domain-independent data flows and data quality rules can likely be generated by AI agents~\cite{akella2025codegenwrangler}, domain-specific data quality is likely to remain a challenge because training for AI agents rarely include these, or if they do, there is a long tail. The same issues exist with anomaly detection; all of which are opportunities for future work. 

\subsection{Monitoring and Remediating Data Flows}
Following the deployment of data flow pipelines, continuous monitoring and management are essential to ensure the reliable flow of data and timely detection of issues. Traditional methods rely on instrumenting dataflow code for observability, monitoring dashboards and alerts, and manual diagnostics, which are often time-intensive and prone to delays. Remediation typically requires domain-specific expertise and coordination among multiple stakeholders, making the process inefficient and, in complex scenarios, potentially stretching from hours to several weeks.

AI agents can integrate with existing tools to monitor data flows and quality, detect anomalies, generate and apply data remediations. Data lineage tracking software already exists and can be useful context for resolving data flow and quality issues~\cite{manta}. Data preparation for report generation and processing provenance for compliance are all candidates for agentic AI. Repairing SQL pipelines, for instance, is an area of active research~\cite{li2025swe}.

\textbf{AI Challenges:}
Data flows can generate massive volumes of heterogeneous data—varying in source, type, and format—which remain challenging for current AI agents to handle efficiently. Scaling these agents to autonomously process, analyze, and remediate such data flows—including generating scripts and code for corrective actions—is an ongoing challenge and an active area of research. This is further compounded by the need to ensure that automated changes adhere to organizational policies and do not introduce unintended harm. Managing cascading effects, particularly in critical operations such as data modifications, schema updates, or frequently executed scripts, often requires human oversight and careful validation to maintain trust and system integrity. Future research is needed to develop autonomous agents capable of operating without such oversight and validation.

\subsection{Ensuring Data Governance and Compliance}
Governance is enforced through manual policy definition and implementation. This includes access control, audit logging, and compliance with regulations.

Agents can classify data (for example, identifying PII), enforce access policies dynamically, and generate audit logs. They can generate pipelines or quality checks in compliance with business process specifications, which map standards like GDPR, HIPPA or U.S. Securities and Exchange Commission (SEC) regulations to enterprise specific directives. Leveraging business process specifications as context for generating and adapting data policies and pipelines enables the system to adapt as the policy and data landscape evolves.  Related work in this space uses LLMs to identify sensitive data, and de-identify it \cite{10.1145/3714334.3714380}, and to audit database access control \cite{subramaniam2025deploiapplyingnl2sqlsynthesize}.

\textbf{AI Challenges:}
Enforcing governance is complex; most regulation is buried in multiple legal documents, and contains complex requirements on storage, use of data, as well as where data can be moved to.  Challenges exist both in automatically generating business process specifications from regulatory documents, and in generating dataflow pipelines and quality rules to implement those business processes. As with anomaly detection and data quality, AI techniques need significant work to become helpful in this space.

\subsection{Generating Actionable Insights}
Enterprises are driven by reports and decision generation, to summarize key informational needs~\cite{ding2023insightpilot,data_story} and to generate potential actions to benefit customers.  Data scientists manually build models, dashboards, and reports. This process is iterative and resource-intensive. Agents can automate model training and evaluation, generate dashboards using tools like Power BI, and summarize insights in natural language~\cite{ding2019quickinsights,ding2023insightpilot}. They can also personalize insights for different user roles and deliver them as new data arrives.

\textbf{AI Challenges:}
While AI has made a lot of strides in the space of automated model building~\cite{SALEHIN202452} across modalities (e.g., for timeseries~\cite{garza2025timecopilot}) and to some extent on building even more complex Data Science pipelines (e.g., ~\cite{nam2025dsstardatascienceagent, hong-etal-2025-data}), a key opportunity lies in determining from a set of data what the key insights actually are, where data might be missing to create the key insights for a given domain, and where one might acquire such data. Another AI challenge area is tracking or identifying that actionable insights have been actioned. 
\newline

The preceding subsections focus on individual layers or tasks within the data stack. However, autonomous orchestration across the entire data stack presents additional challenges and opportunities. In the following section, we focus on foundational cross-stack capabilities required to build this autonomous system.

%% file: body/vision_new.tex
Realizing a fully autonomous data stack presents challenges at both the task level, as discussed in Section~\ref{sec:datastack}, and in orchestrating automations across the entire stack, as discussed in this section. Although some of the agentic approaches at the task level are already being developed, we are unaware of prior work on fully autonomous data stacks. For this section, we focus on foundational capabilities we believe must underpin a fully autonomous data stack and highlight research opportunities. We anticipate task-level automations can be developed and deployed incrementally to bring scale, efficiency and reliability in the near term, while the more overarching challenges in this section will enable the community to pursue even greater levels of autonomy.

\subsection{Continuous Learning via Feedback and Observability}
Continuous feedback signals and end-to-end observability is essential to build autonomous agents that adapt and improve over time in a self-supervised manner. 
These signals must be used properly and must propagate through all layers of the agentic system to ensure better outcomes. These signals at times, can lead to different types of actions, for instance, they can source higher-quality data to address the quality of the output, or reconfigure infrastructure to address latency or accuracy issues.

While useful, collecting and managing such feedback can be significantly challenging, and  require additional infrastructure, including ingestion pipelines or storage. This challenge is further compounded when the root cause of an issue lies in a layer different than where it manifested, necessitating the need for a comprehensive observability stack. While some tools like Instana, Turbonomics, Dynatrace, DataDog, Grafana can offer partial insights, a comprehensive observability stack across all layers of the data stack from user interface to data storage is essential for effective and automated diagnosis, remediation, and continuous improvement. However, just as agents can tackle autonomous actions for certain tasks or layers without requiring the full stack to be autonomous, we need not wait for a full observability solution to derive value.

\subsection{Providing context for data agent grounding}
AI agents require sufficient context when performing a task to understand the nuances of the task, connect information, generate relevant and accurate responses, and avoid "hallucinations" or nonsensical outputs. Data engineers similarly require context in performing data management and operations tasks.  For example, a business glossary is often used to label tables and columns with enterprise standard terminology and semantic meaning. Data compliance regulations and business processes are the rules and guidelines for how a business should collect, store, process, and protect data. Thus, these rules and guidelines provide context for data engineers to configure data systems and implement data processing pipelines to ensure compliance.

A recent move towards open standards for specifying a number of data management and operations artifacts may present an opportunity to ground data agents with necessary context. For example, data products are an emerging approach to creating structured, reusable, and well-defined data assets. Open standards are emerging for data products, which specify objects, attributes, and structure \cite{ODPS_v4_0}; data product contracts, which specify schema, data quality, service levels and other properties \cite{ODCS2025}; and consistent collection of lineage metadata describing how a data product is produced and used \cite{OpenLineage}. While we are not advocating any standard in particular, we note that such specifications could be instrumental in how agents go from natural language requests to machine-readable specifications of context, requirements, and definitions to help guide agent collaboration. Additional research is required into grounding data agents and their orchestration with appropriate business processes and data stack context.


\subsection{Autonomous Planning and Orchestration}
\begin{figure*}
    \centering
    \includegraphics[scale=0.85,trim={0 6cm 11cm 3cm},clip]{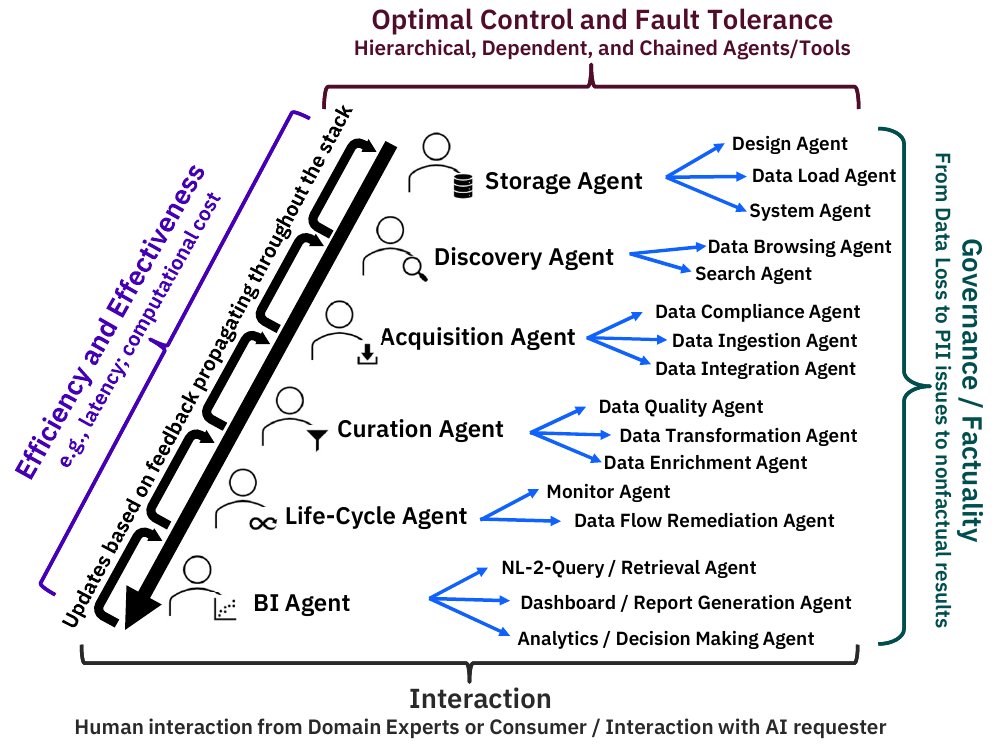}
\caption{Architecture of an Agentic DataOps system showing hierarchical layers of the data stack and corresponding specialized agents. The figure illustrates that optimal control, fault tolerance, and continuous feedback propagation through observability are needed for an autonomously deployed system. Controlling such an autonomous stack is challenging due to the sheer scale and interdependence of tasks, diverse requirements for planning, and critical constraints such as efficiency, governance, factuality, and external steerability—whether by humans or other AI systems. Note that the shown list of agents is not complete and that an agent might not just call tools, but invoke additional agents as well.}
\label{orchestration}
\end{figure*}

A fundamental capability of modern large language model (LLM)-based agentic systems is their ability to autonomously plan, adapt, and orchestrate complex tasks through the integration of diverse tools. Given a high-level objective in natural language, these systems can interpret user intent, formulate multi-step plans, and execute them by invoking appropriate tools in sequence. 
As shown in Figure~\ref{orchestration}, building and controlling a fully autonomous data stack requires more than executing isolated tasks—it demands coordinated planning across interdependent layers. The system must identify relevant tasks (e.g., infrastructure design, data discovery, enrichment, quality assessment, and transformation), select appropriate agents and tools, and determine execution order under constraints such as computational cost, governance, and factuality, the latter being particularly critical given the tendency of AI models to produce nonfactual outputs.
Beyond execution, the architecture incorporates observability for debugging and continuous feedback propagation to, for instance, optimize efficiency and effectiveness. It must also address dynamic issues, such as reconfiguring infrastructure to meet latency requirements, while remaining steerable by external actors—either human experts or other AI systems—to ensure adaptability and trust.


Recent work such as InsightPilot~\cite{ding2023insightpilot} and Spider 2.0~\cite{lei2024spider2} demonstrates how agents can be orchestrated to perform complex data exploration and query generation tasks in challenging data settings. Similarly, DocETL~\cite{shankar2024docetl} introduces agentic query rewriting for document-centric pipelines. These systems are examples for how automated planning in scoped tasks can be grounded in context and adapted to evolving data and user needs.
Crucially, the planning process is dynamic—adjusting in real time based on the outcomes of preceding steps—and includes mechanisms to engage the user when ambiguities or uncertainties arise. For example, in a financial asset prediction scenario, an agent may initially identify relevant datasets; based on an assessment of their quality and completeness, it can modify its strategy to locate alternative data sources, cleanse existing data, or initiate a request to a data steward for additional inputs. The extensibility of these systems—through the integration of new tools—and their capacity to incorporate human-in-the-loop feedback are essential for enabling flexible, context-aware, and robust task execution, making them a cornerstone of future intelligent and autonomous systems.


\subsection{Human Supervision, Auditability, and Secure Autonomy}
A critical challenge is ensuring agents do not operate beyond intended boundaries. Robust mechanisms for integrating human oversight and intervention are required. Rigorous protocols should direct when and how these agents should request human intervention, especially in ambiguous and risky scenarios. In addition, the systems must also support robust fail-safe and deactivation mechanisms to allow graceful overrides or shutdowns when agents behave unpredictably or unsafely.

Fully autonomous agents must support native auditability, including their internal decision-making process. Such audit mechanisms are critical for backward traceability, retrospective analysis, regulatory compliance, debugging, and performance optimization. As agents become more autonomous, their security risk also increases. These agents will require policy-based access control, usage monitoring, and automated compliance enforcement. Agents should also support real-time risk assessment, detect behavioral anomalies, and maintain secure, tamper-evident logs. These features are essential for protecting sensitive data and enabling autonomous agents to align with organizational policies, legal requirements, and user expectations.


\subsection{Multi-Agent Collaboration and Communication}
The proposed data stack would require a multitude of agents, developed by diverse teams, individuals, and even by different organizations. Given this inherent heterogeneous nature of agents, it becomes critical that these agents interoperate seamlessly towards a shared goal. Efficient and effective approaches for agents to exchange context, intentions, and actions, is key multi-disciplinary research topic. 
Several technologies have started to emerge, indicating early progress in this direction. A2A (Agent-to-Agent Communication Protocol) and MCP (Model Context Protocol) are some examples of such efforts. Further research is needed to enable effective multi-agent collaboration and communications principles and mechanisms to enable building an autonomous data stack.

\subsection{Benchmarking Agentic Systems \& Simulations} 
Any automation is incomplete without a comprehensive evaluation of its effectiveness across different dimensions, ranging from technical capabilities to user experience. 
While there has been some progress in this area, i.e., some benchmarks exist (e.g.,~\cite{ELT-Bench, sembench, liu2023agentbench, li2024can, BirdInteract, xu2025systemsfoundationsagenticexploration, tbench_2025, fdabench, SQLStorm}), primarily focusing on individual narrow tasks such as Text-to-SQL with or without semantic operators, ETL flow creation, analytics over heterogeneous data, or operations in terminal environments, comprehensive standardized frameworks and benchmarks along with evaluation methods that can assess the system end-to-end across different dimensions, including business key performance indicators (KPIs), remain an open challenge and a future area of research. 
\begin{table}[ht]
\centering
\begin{tabular}{|p{1.5cm}|p{2.9cm}|p{3.5cm}|}
\hline
\textbf{Dimension} & \textbf{KPI} & \textbf{Description} \\
\hline
Latency & Task Completion Time & Time taken from request to insight generation \\
\hline
System Cost & Computational cost & CPU, GPU, network and storage cost \\
\hline
Service Cost & Cost for services & Acquisition and subscription cost for data, etc. \\
\hline
Autonomy & Human Intervention Rate & Percentage of tasks completed without human input \\
\hline
Accuracy & Insight Precision \& Recall & Correctness of generated insights or models \\
\hline
Adaptability & Feedback Responsiveness & Speed and effectiveness of system adaptation to feedback \\
\hline
Compliance & Policy Violation Rate & Number of governance or regulatory violations detected \\
\hline
Scalability & Throughput & Number of concurrent data tasks handled without degradation \\
\hline
Robustness & Fault Recovery Time & Time to detect and remediate pipeline or system failures \\
\hline
Explainability & Auditability Score & Ability to trace agent decisions and actions retrospectively \\
\hline
\end{tabular}
\caption{High-level KPIs proposed for benchmarking autonomous data systems, capturing critical dimensions required to comprehensively evaluate system performance across efficiency, reliability, scalability, and compliance.}
\label{tab:kpi-agentic}
\end{table}
Table~\ref{tab:kpi-agentic} summarizes KPIs for benchmarking agentic data systems across multiple dimensions. These include operational metrics such as latency (task completion time) and cost (computational resources as well as data services), autonomy (human intervention rate), and accuracy (precision and recall of insights). Adaptability is measured through feedback responsiveness, while compliance focuses on policy violation rates. Scalability and robustness are captured by throughput and fault recovery time, respectively, and explainability is assessed via auditability scores. Together, these KPIs provide a comprehensive framework for evaluating efficiency, reliability, and trustworthiness in autonomous data stacks.

The Agent Company~\cite{agentcompany} takes an important step in this direction by tracking the performance of LLM-based agents on real-world professional tasks by providing an extensible benchmark. This benchmark evaluates agents that operate like digital workers—browsing the web, writing code, executing programs, and collaborating with other agents or users.
Furthermore, simulation imitating the real world scenario is another area of research to ensure that these evaluations are not static, and capture the real world setting so that any changes made to the system only lead to their improvement without any unwanted outcome.

\subsection{Community Participation}
Previous subsections have focused on research and technical challenges.  However, we also believe that a critical underpinning to the future of autonomous data systems will be the active participation of the data systems community itself. That is, opportunities exist not only in the AI required, but also in the underlying data systems and architecture, such as designing core components across the entire stack for autonomy. We believe the community should come together to brainstorm how we might better tackle such an ambitious agenda. The following are examples of where community participation could foster collaboration and progress.
\begin{enumerate}
\item Building a comprehensive taxonomy for tasks in each layer of data stack so that corresponding tools can be built. 
 \item Open Standards for End to End Observability: Unified monitoring data by expanding existing standard like OTEL to accommodate data flow specific monitoring data
\item Release of Benchmark including Data Flows, Fault Injection system, corresponding observability data and framework to plug remediation agents.
\item Course Curriculum to extend Data Management Systems to include designing agents and subsystems for autonomous operations.
\item Development of focused use cases (grounded in real world applications) to show case the utility of Agentic DataOps.
\end{enumerate}


%% file: body/data_stack_opportunities.tex

The adoption of AI-powered autonomous data stacks would bring transformative changes to enterprise data management and operations, well beyond the efficiencies gained and the roles and responsibilities of those who manage enterprise data estates today. In this section, we consider the opportunities the future of autonomous data systems might offer.

\begin{enumerate}
   \item \textbf{Natural Language to Business} Due to the complexities of gathering, curating, and serving data, many businesses specialize in data-centric services.  For example, data curation companies specialize in preparing datasets by labeling, annotating, and resolving quality issues. Data enrichment companies enhance existing datasets by adding valuable information from external sources, such as demographics, firmographics, or behavioral data. In the spirit of a recent experiment\footnote{\url{https://www.anthropic.com/research/project-vend-1}} on AI running a small business, might autonomous data stacks make this possible for data-centric businesses?

   \item \textbf{Fully Empowered Knowledge Workers} Natural language interfaces to any data would enable a broader range of stakeholders—regardless of technical background—to interact meaningfully with data. Business users, product teams, and domain experts would be able to specify analytical goals, evaluate results, and iterate without relying on technical intermediaries. The required human expertise might transition further toward strategic decision-making, ethical judgment, and effectively guiding agents towards the most valuable collection of data, and its refinement into information,  knowledge, and insights. 


    \item \textbf{AI for AI} The autonomous data stack enables AI systems to independently resolve complex albeit specific tasks — tasks that may themselves be generated by another AI. For example, consider our ongoing stock trading scenario: this could be just one sub-task within a broader AI-driven wealth management service. Such a system might also encompass investment strategies across real estate, commodities, and other asset classes, all orchestrated by AI.

    \item \textbf{Data and Tool Ecosystems}
    Just as smartphones enabled the rise of app ecosystems, autonomous data agents might lead to the emergence of data and tool ecosystems. Such ecosystems might be tailored to industry-specific needs or other paradigms. These ecosystems could facilitate reusable workflows, shared data tools, and even cross-organizational knowledge sharing. 



    \item \textbf{Continuous Data Compliance}
    Today, audits and other compliance reviews incur enormous manual efforts and, in turn, expense.  As a result, enterprise data audits are typically conducted once or twice a year. Technical staff must act upon the results of data lineage traces, painstakingly resolving any non-compliance issues.  Business process specialists and compliance officers must monitor new regulations and standards, and transform into business process directives and specifications, which in turn require updates to data systems and workflows. Could an autonomous data stack continuously monitor compliance of the entire enterprise data estate and resolve issues as they arise?  Could it enable real-time adherence to evolving regulations and standards?
    
\end{enumerate}



%% file: body/conclusion.tex

In this paper, we proposed the vision of Agentic DataOps that aims to reimagine the data stack with dedicated agents for autonomy at each layer and, at the same time, holistically across all layers. While a fully autonomous data stack is the vision, we emphasized the need for making incremental progress in each of the sublayers by building better tools and intelligent agents. Such a transformational pivot could not happen without broader community participation, and we called out areas where the data systems research community might collaborate. Finally, we believe that the vision, while ambitious, could lead to enormous opportunities, which we also hypothesized. We believe that many in the data systems community who are also working on AI-infused data layers, are likewise envisioning a more autonomous future. We hope that by offering this vision to the community, we can better join forces.


%% file: body/acknowledgements.tex
We sincerely appreciate the contributions of our colleagues including William Maness, Edward Calvesbert, Enzo Cialini.